\pdfoutput=1

\documentclass[11pt]{article}

\usepackage[affil-it]{authblk}

\usepackage{acl}

\usepackage{times}
\usepackage{latexsym}

\usepackage[T1]{fontenc}

\usepackage[utf8]{inputenc}

\usepackage{microtype}

\usepackage{graphicx,caption}
\usepackage{booktabs}
\usepackage{multicol,multirow}
\usepackage{amsmath,amsfonts,amssymb}
\usepackage{bm}

\title{Beyond the Granularity: Multi-Perspective Dialogue Collaborative Selection for Dialogue State Tracking}

\author[1]{\textbf{Jinyu Guo}}
\author[ ]{\textbf{Kai Shuang}\textsuperscript{1}\thanks{\; Corresponding author.}}
\author[1]{\textbf{Jijie Li}}
\author[2]{\textbf{Zihan Wang}}
\author[1]{\textbf{Yixuan Liu}}
\affil[1]{State Key Laboratory of Networking and Switching Technology,}
\affil[ ]{Beijing University of Posts and Telecommunications}
\affil[2]{Graduate School of Information Science and Technology, The University of Tokyo}
\affil[ ]{\texttt{\{guojinyu, shuangk, lijijie, liuyixuan\}@bupt.edu.cn}}
\affil[ ]{\texttt{zwang@tkl.iis.u-tokyo.ac.jp}}

\date{}

\begin{document}
\maketitle
\begin{abstract}
In dialogue state tracking, dialogue history is a crucial material, and its utilization varies between different models. However, no matter how the dialogue history is used, each existing model uses its own consistent dialogue history during the entire state tracking process, regardless of which slot is updated. Apparently, it requires different dialogue history to update different slots in different turns. Therefore, using consistent dialogue contents may lead to insufficient or redundant information for different slots, which affects the overall performance. To address this problem, we devise DiCoS-DST to dynamically select the relevant dialogue contents corresponding to each slot for state updating. Specifically, it first retrieves turn-level utterances of dialogue history and evaluates their relevance to the slot from a combination of three perspectives: (1) its explicit connection to the slot name; (2) its relevance to the current turn dialogue; (3) Implicit Mention Oriented Reasoning. Then these perspectives are combined to yield a decision, and only the selected dialogue contents are fed into State Generator, which explicitly minimizes the distracting information passed to the downstream state prediction. Experimental results show that our approach achieves new state-of-the-art performance on MultiWOZ 2.1 and MultiWOZ 2.2, and achieves superior performance on multiple mainstream benchmark datasets (including Sim-M, Sim-R, and DSTC2).\footnote{Code is available at \\https://github.com/guojinyu88/DiCoS-master}
\end{abstract}

\begin{figure}[t]
\centering
\includegraphics[width=0.48\textwidth]{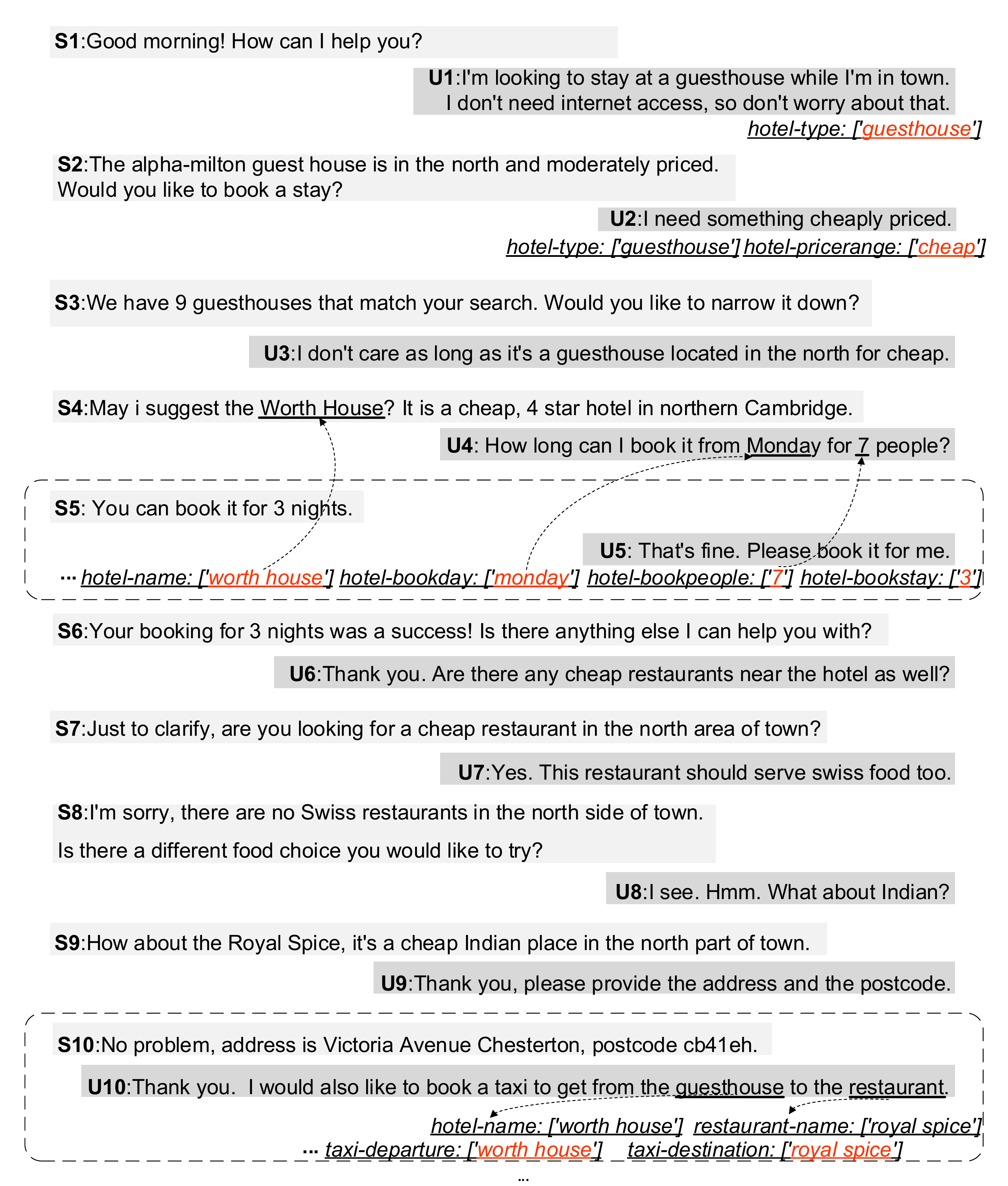}
\caption{An example of multi-domain dialogues. Utterances at the left and the right sides are from system and user, respectively. Each red slot value in the figure indicates that it is updated in its turn.}
\label{fig:example}
\end{figure}

\section{Introduction}

Task-oriented dialogue systems have recently attracted growing attention and achieved substantial progress. Dialogue state tracking (DST) is a core component, where it is responsible for interpreting user goals and intents and feeding downstream policy learning in dialogue management. The common practice treats it as a problem of compacting the dialogue content into a series of slot-value pairs that represent information about the user goals updated until the current turn. For example, in Figure~\ref{fig:example}, the dialogue state at turn 2 is \{(``$hotel-type$'', ``$guesthouse$''), (``$hotel-pricerange$'', ``$cheap$'')\}.

In dialogue state tracking, dialogue history is a crucial source material. Recently, granularity has been proposed to quantify the utilization of dialogue history\cite{yang-etal-2021-comprehensive}. In DST, the definition of granularity is the number of dialogue turns spanning from a certain dialogue state in the dialogue to the current dialogue state. Traditional DST models usually determine dialogue states by considering only utterances at the current turn (i.e., $\mathrm{granularity}=1$), while recent researches attempt to utilize partial history (i.e., $\mathrm{granularity}=k,\ k<T$) or introduce all dialogue history information into the prediction (i.e., $\mathrm{granularity}=T$). However, no matter what granularity is used, we find that each model uses a constant granularity it determines, regardless of which slot is being updated. Apparently, it requires different granularity for different slots in different turns. For example, in Figure~\ref{fig:example}, the granularity required for slot ``$hotel-name$'', ``$hotel-bookday$'', and ``$hotel-bookpeople$'' in turn 5 is 2, while slot ``$hotel-bookstay$'' in turn 5 requires a granularity of 1. Therefore, using a constant granularity may lead to insufficient input for updating some slots, while for others, redundant while confusing contents can become distracting information to pose a hindrance, which affects the overall performance.

Furtherly, granularity means directly working on all dialogue contents from a particular turn to the current turn, regardless of the fact that there are still dialogue contents that are not relevant to the slot. Therefore, if it is possible to break the limitation of granularity and to dynamically select relevant dialogue contents corresponding to each slot, the selected dialogue contents as input will explicitly minimize distracting information being passed to the downstream state prediction.

To achieve this goal, we propose a DiCoS-DST to fully exploit the utterances and elaborately select the relevant dialogue contents corresponding to each slot for state updating. Specifically, we retrieve turn-level utterances of dialogue history and evaluate their relevance to the slot from a combination of three perspectives. First, we devise an SN-DH module to touch on the relation of the dialogue and the slot name, which straightforward reflects the relevance. Second, we propose a CT-DH module to explore the dependency between each turn in the dialogue history and the current turn dialogue. The intuition behind this design is that the current turn dialogue is crucial. If any previous turn is strongly related to the current turn dialogue, it can be considered useful as dependency information for slot updating. Third, we propose an Implicit Mention Oriented Reasoning module to tackle the implicit mention (i.e., coreferences) problem that commonly exists in complex dialogues. Specifically, we build a novel graph neural network (GNN) to explicitly facilitate reasoning over the turns of dialogue and all slot-value pairs for better exploitation of the coreferential relation information. After the evaluation of these three modules, we leverage a gate mechanism to combine these perspectives and yield a decision. Finally, the selected dialogue contents are fed into State Generator to enhance their interaction, form a new contextualized sequence representation, and generate a value using a hybrid method.

We evaluate the effectiveness of our model on most mainstream benchmark datasets on task-oriented dialogue. Experimental results show that our proposed DiCoS-DST achieves new state-of-the-art performance on both two versions of the most actively studied dataset: MultiWOZ 2.1~\cite{eric2019multiwoz} and MultiWOZ 2.2~\cite{zang2020multiwoz} with joint goal accuracy of 61.02\% and 61.13\%. In particular, the joint goal accuracy on MultiWOZ 2.2 outperforms the previous state-of-the-art by 3.09\%. In addition, DiCoS-DST also achieves new state-of-the-art performance on Sim-M and Sim-R~\cite{shah2018building} and competitive performance on DSTC2~\cite{henderson2014second}.

Our contributions in this work are three folds:
\begin{itemize}
    \item We propose a Multi-Perspective Dialogue Collaborative Selector module to dynamically select relevant dialogue contents corresponding to each slot from a combination of three perspectives. This module can explicitly filter the distracting information being passed to the downstream state prediction.
    \item We propose Implicit Mention Oriented Reasoning and implement it by building a GNN to explicitly facilitate reasoning and exploit the coreferential relation information in complex dialogues.
    \item Our DiCoS-DST model achieves new state-of-the-art performance on the MultiWOZ 2.1, MultiWOZ 2.2, Sim-M, and Sim-R datasets.
\end{itemize}

\begin{figure*}[t]
\centering
\includegraphics[width=\textwidth]{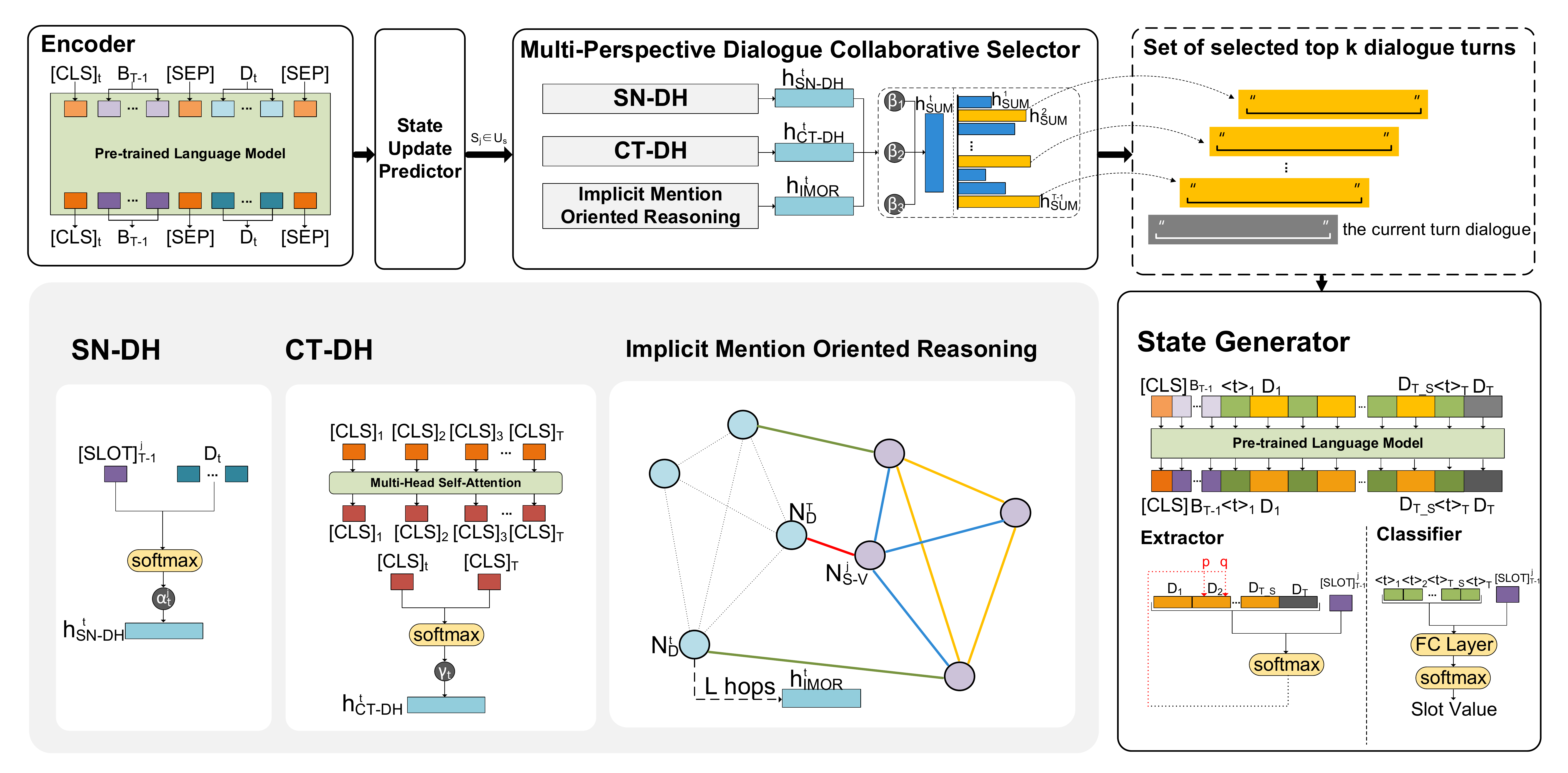}
\caption{The architecture of the proposed DiCoS-DST model. The gray area in the lower left part of the figure shows the internal structure of the three modules in Multi-Perspective Dialogue Collaborative Selector.}
\label{fig:framework}
\end{figure*}

\section{Related Work}

There has been a plethora of research on dialogue state tracking. Traditional dialogue state trackers relied on a separate Spoken Language Understanding (SLU) module~\cite{thomson2010bayesian,wang2013simple} to extract relevant information. In recent years, neural network models are proposed for further improvements. One way to classify DST models is whether they use dialogue history. Some DST models obtain each slot value in the dialogue state by inquiring about a part or all of the dialogue history~\cite{xu2018end,lei2018sequicity,goel2019hyst,ren2019scalable,shan2020contextual,zhang2020find,chen2020schema,guo2021dual}, while the others use the current turn dialogue to predict the dialogue state~\cite{mrkvsic2017neural,kim2020efficient,heck2020trippy,zhu2020efficient}. Recently, \cite{yang-etal-2021-comprehensive} first proposed the granularity in DST to quantify the use of dialogue history. Its experimental results show that different models on different datasets have different optimal granularity (not always using the entire dialogue history). However, no matter what granularity is used, each model uses a constant granularity it determines, regardless of which slot is updated.

On the other hand, dialogue state tracking and machine reading comprehension (MRC) have similarities in many aspects~\cite{gao2020machine}. Recently, Multi-hop Reading Comprehension (MHRC) has been a challenging topic. For cases in MHRC datasets, one question is usually provided with several lexically related paragraphs, which contain many confusing contexts. To deal with this situation, cascaded models~\cite{qiu2019dynamically,groeneveld2020simple,tu2020select,wu2021graph} that are composed of a reader and a retriever are often used. They retrieve the most relevant evidence paragraphs first and perform multi-hop reasoning on retrieved contexts thereafter. The mechanism of dialogue selection before state generation in our work is partially inspired by the paragraph retrieval in multi-hop reading comprehension.

\section{Approach}

The architecture of DiCoS-DST is illustrated in Figure~\ref{fig:framework}. DiCoS-DST consists of Encoder, State Update Predictor, Multi-Perspective Dialogue Collaborative Selector, and State Generator. Here we first define the problem setting in our work. We define the number of the current turn as $T$. The task is to predict the dialogue state at each turn $t\ (t\le T)$, which is defined as $\mathcal{B}_t=\{(S^j,V_t^j)|1\le j\le J\}$, where $S^j$ is the slot name, $V_t^j$ is the corresponding slot value, and $J$ is the total number of slots. For the sake of simplicity, we omit the superscript $T$ in the variables in the next sections.

\subsection{Encoder}

We employ the representation of the previous turn dialogue state $B_{T-1}$ concatenated to the representation of each turn dialogue utterances $D_t$ as input: $E_t=[\mathrm{CLS}]_t\oplus B_{T-1}\oplus [\mathrm{SEP}]\oplus D_t,\ (1\le t\le T)$, where $[\mathrm{CLS}]_t$ is a special token added in front of every turn input. The representation of the previous turn dialogue state is $B_{T-1}=B_{T-1}^1\oplus\ldots\oplus B_{T-1}^J$. The representation of each slot's state $B_{T-1}^j=[\mathrm{SLOT}]_{T-1}^j\oplus S_j\oplus [\mathrm{VALUE}]_{T-1}^j\oplus V_{T-1}^j$, where $[\mathrm{SLOT}]_{T-1}^j$ and $[\mathrm{VALUE}]_{T-1}^j$ are special tokens that represent the slot name and the slot value at turn $T-1$, respectively. We donate the representation of the dialogue at turn $t$ as $D_t=R_t\oplus;\oplus U_t\oplus[\mathrm{SEP}]$, where $R_t$ is the system response and $U_t$ is the user utterance. $;$ is a special token used to mark the boundary between $R_t$ and $U_t$, and $[\mathrm{SEP}]$ is a special token used to mark the end of a dialogue turn. 

Then a pre-trained language model (PrLM) will be adopted to obtain contextualized representation for the concatenated input sequence $E_t$.

\subsection{State Update Predictor}

We attach a two-way classification module to the top of the Encoder output. It predicts which slots require to be updated in the current turn. The subsequent modules will only process the selected slots, while the other slots will directly inherit the slot values from the previous turn.

We inject this module because whether a slot requires to be updated indicates whether the current turn dialogue is significant for this slot. For CT-DH of the subsequent Multi-Perspective Collaborative Selector, the great importance of the current turn dialogue is a prerequisite. A more detailed explanation will be given in Section~\ref{sec:multiperspective_collaborative_selector}.

We employ the same mechanism as \cite{guo2021dual} to train the module and to predict the state operation. We sketch the prediction process as follows:
\begin{equation}
    \mathrm{SUP}(S_j)=\left\{
        \begin{array}{rl}
            \mathrm{update}, & \mathrm{if}\ \mathrm{Total}\_score_j>\delta \\
            \mathrm{inherit}, & \mathrm{otherwise}
        \end{array}
    \right.
\end{equation}

We define the set of the selected slot indices as $\bm{\mathrm{U}}_s=\{j|\mathrm{SUP}(S_j)=\mathrm{update}\}$.

\subsection{Multi-Perspective Dialogue Collaborative Selector}
\label{sec:multiperspective_collaborative_selector}

For each slot $S_j\ (j\in \bm{\mathrm{U}}_s)$ selected to be updated, SN-DH, CT-DH, and Implicit Mention Oriented Reasoning modules are proposed to evaluate dialogue relevance and aggregate representations from three perspectives. Then a gated fusion mechanism is implemented to perform the dialogue selection.

\paragraph{SN-DH}
SN-DH (Slot Name - Dialogue History) aims to explore the correlation between slot names and each turn of the dialogue history. For slot $S_j$, the slot name is straightforward explicit information. Therefore, the correlation with the slot name directly reflects the importance of the dialogue turn. We take the slot name presentation $[\mathrm{SLOT}]_{T-1}^j$ as the attention to the $t$-th turn dialogue representation $D_t$. The output $\bm{\alpha}_t^j=\mathrm{softmax}(D_t([\mathrm{SLOT}]_{T-1}^j)^\intercal)$ represents the correlation between each position of $D_t$ and the $j$-th slot name at turn $t$. Then we get the aggregated dialogue representation $h_\mathrm{SN-DH}^t=(\bm{\alpha}_t^j)^\intercal D_t$, which will participate in the subsequent fusion as the embedding of the $t$-th turn dialogue in this perspective.

\paragraph{CT-DH}
As aforementioned, a slot that needs to be updated in the current turn means that the current turn dialogue is most relevant to this slot. In this case, if the dialogue content of any other turn contains the information that the current turn dialogue highly depends on, it can also be considered useful. Based on this consideration, we devise a CT-DH (Current Turn - Dialogue History) module to explore this association. Specifically, we build a multi-head self-attention (MHSA) layer on top of the $[\mathrm{CLS}]$ tokens generated from different turns of dialogue to enhance inter-turn interaction. The MHSA layer is defined as:
\begin{gather}
    head_i=\mathrm{Attention}(QW_i^Q,KW_i^K,VW_i^V) \\
    Multihead=(head_i\oplus\ldots\oplus head_n)W^O \\
    I=\mathrm{MHSA}([\mathrm{CLS}]_1\oplus\ldots\oplus [\mathrm{CLS}]_T)
\end{gather}
where $Q$, $K$, and $V$ are linear projections from $[\mathrm{CLS}]$ embeddings of each turn of dialogue, representing attention queries, key and values.

We then append an attention layer between the output representation of the current turn dialogue and each turn of dialogue history to capture interactions between them:
\begin{gather}
    \bm{\gamma}_t=\mathrm{Attention}([\mathrm{CLS}]_t,[\mathrm{CLS}]_T) \\
    h_\mathrm{CT-DH}^t=\bm{\gamma}_t[\mathrm{CLS}]_T+[\mathrm{CLS}]_t
\end{gather}

$h_\mathrm{CT-DH}^t$ will participate in the subsequent fusion as an aggregated representation of the $t$-th dialogue in this perspective.

\begin{figure}[t]
\centering
\includegraphics[width=0.4\textwidth]{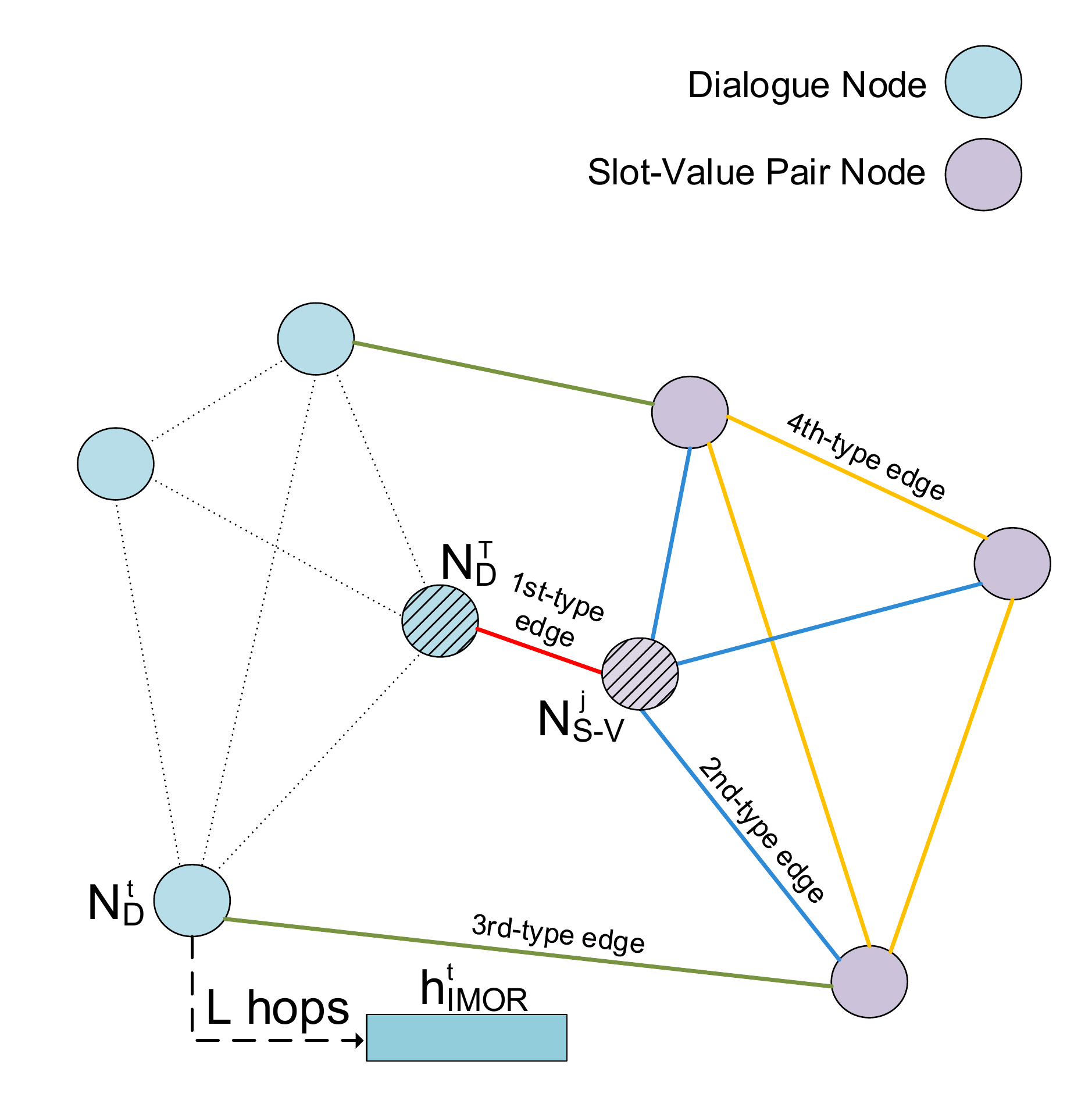}
\caption{Diagram of the graph neural network. The dashed connection between the dialogue nodes does not actually exist. We draw them to show that using the dialogue representation output by MHSA already includes the contextual interactions between the dialogues.}
\label{fig:graph}
\end{figure}

\paragraph{Implicit Mention Oriented Reasoning}
Handling a complex dialogue usually requires addressing implicit mentions (i.e., coreferences). As shown in Figure~\ref{fig:example}, in turn 10, the restaurant is not referred to explicitly upon ordering a taxi within the same dialogue turn. Instead, it is present in the value of another slot. Therefore, SN-DH and CT-DH are difficult to deal with this case due to their mechanisms. To tackle this problem, we build a graph neural network (GNN) model to explicitly facilitate reasoning over the turns of dialogue and all slot-value pairs for better exploitation of the coreferential relation. As illustrated in Figure~\ref{fig:graph}, the nodes in the graph include two types: $N_D$ for each turn dialogue and $N_{S-V}$ for each slot-value pair. They are initialized with the MHSA output representation $[\mathrm{CLS}]_t$ and $W_{S-V}([\mathrm{SLOT}]_{T-1}^z\oplus [\mathrm{VALUE}]_{T-1}^z)\ (1\le z\le J)$, respectively. Then we design four types of edges to build the connections among graph nodes: 
\\1) Add an edge between $N_{S-V}^j$ and $N_D^T$ (red line in Figure~\ref{fig:graph}). As aforementioned, the slot $S_j$ will be updated. This edge is to establish the connection between the slot to be updated and the current turn dialogue; 
\\2) Add an edge between $N_{S-V}^j$ and $N_{S-V}^z\ (z\neq j)$ (blue line in Figure~\ref{fig:graph}). These edges are to establish connections between the slot to be updated and other slots; 
\\3) Add an edge between $N_{S-V}^z\ (z\neq j)$ and $N_D^{t_z}$. $t_z$ is the turn when the most up-to-date value of 
$S_z$ is updated (green line in Figure~\ref{fig:graph}). These edges are to establish connections between each slot and the turn of dialogue in which its latest slot value was updated; 
\\4) Add an edge between $N_{S-V}^{z_1}$ and $N_{S-V}^{z_2}$ ($S_{z_1}$ and $S_{z_2}$ belong to the same domain) (yellow line in Figure~\ref{fig:graph}). These edges are to establish connections between slots that belong to the same domain.

The motivation for this design is that we first explore the relation between the slot to be updated and other slot-value pairs based on the current turn dialogue. Then we use other slot-value pairs as media to establish relations to their corresponding dialogue turns. We add the fourth type of edges to represent the auxiliary relationship of slots that belong to the same domain.

We use multi-relational GCN with gating mechanism as in \cite{de2019question,tu2019multi}. We define $h_i^0$ represents initial node embedding from $N_D$ or $N_{S-V}$. The calculation of node embedding after one hop can be formulated as:
\begin{gather}
    h_i^{l+1}=\sigma(u_i^l)\odot g_i^l+h_i^l\odot (1-g_i^l) \\
    u_i^l=f_s(h_i^l)+\sum_{r\in \mathcal{R}}\frac{1}{|\mathcal{N}_i^r|}\sum_{n\in \mathcal{N}_i^r}f_r(h_n^l) \\
    g_i^l=\mathrm{sigmoid}(f_g([u_i^l;h_i^l]))
\end{gather}

$\mathcal{N}_i^r$ is the neighbors of node $i$ with edge type $r$, $\mathcal{R}$ is the set of all edge types, and $h_n^l$ is the node representation of node $n$ in layer $l$. $|\cdotp|$ indicates the size of the neighboring set. Each of $f_r$, $f_s$, $f_g$ can be implemented with an MLP. Gate control $g_i^l$ is a vector consisting of values between 0 and 1 to control the amount information from computed update $u_i^l$ or from the original $h_i^l$. Function $\sigma$ denotes a non-linear activation function.

After the message passes on the graph with $L$ hops, we take the final representation of the $t$-th turn dialogue node $N_D^t$ as the aggregated representation $h_\mathrm{IMOR}^t$ in this perspective.

\paragraph{Gating Fusion and Collaborative Selection}
The representations $h_\mathrm{SN-DH}^t$, $h_\mathrm{CT-DH}^t$, and $h_\mathrm{IMOR}^t$ of the $t$-th turn dialogue enter this module for fusion and ranking. To balance the information from multiple perspectives, we leverage a gate mechanism to compute a weight to decide how much information from each perspective should be combined. It is defined as follows:
\begin{gather}
    \beta_1=\sigma_1(W_{\beta_1}\tanh(W_1 h_\mathrm{SN-DH}^t)) \\
    \beta_2=\sigma_2(W_{\beta_2}\tanh(W_2 h_\mathrm{CT-DH}^t)) \\
    \beta_3=\sigma_3(W_{\beta_3}\tanh(W_3 h_\mathrm{IMOR}^t)) \\
    h_\mathrm{sum}^t=\beta_1 h_\mathrm{SN-DH}^t+\beta_2 h_\mathrm{CT-DH}^t+\beta_3 h_\mathrm{IMOR}^t
\end{gather}

After the fusion, an MLP layer is followed, and then we take the dialogues of the top k ranked turns as the selected dialogue contents.

It is worth mentioning that, unlike the state update predictor, since there is no ground-truth label of the dialogue turns that should be selected corresponding to each slot, we take this module and the following state generator as a whole and train it under the supervision of the final dialogue state label. We mark each selected dialogue turn to make the gradient of the state generator losses only backpropagate to the marked turns to ensure the effectiveness of supervision.

\begin{table*}[t]
\centering
\begin{tabular}{l|cc|cc|c|c|c}
\hline
\multicolumn{1}{c|}{\multirow{2}{*}{\textbf{Model}}} & \multicolumn{2}{c|}{\textbf{MultiWOZ 2.1}}                                                                                   & \multicolumn{2}{c|}{\textbf{MultiWOZ 2.2}}                                                                                   & \multicolumn{1}{c|}{\textbf{Sim-M}} & \multicolumn{1}{c|}{\textbf{Sim-R}} & \multicolumn{1}{c}{\textbf{DSTC2}}
\\ \cline{2-8} 
                                & \textbf{\begin{tabular}[c]{@{}c@{}}Joint\end{tabular}} & \textbf{\begin{tabular}[c]{@{}c@{}}Slot\end{tabular}} & \textbf{\begin{tabular}[c]{@{}c@{}}Joint\end{tabular}} & \textbf{\begin{tabular}[c]{@{}c@{}}Slot\end{tabular}} & \textbf{\begin{tabular}[c]{@{}c@{}}Joint\end{tabular}} & \textbf{\begin{tabular}[c]{@{}c@{}}Joint\end{tabular}} & \textbf{\begin{tabular}[c]{@{}c@{}}Joint\end{tabular}} \\ \hline
TRADE & 45.60  & -  & 45.40  & -  & -  & -  & - \\
DST+LU & - & - & - & - & 46.0 & 84.9 & - \\
BERT-DST & - & - & - & - & 80.1 & 89.6 & 69.3 \\
TripPy & 55.29 & - & - & - & 83.5 & 90.0 & - \\
Pegasus-DST & 54.40 & - & 57.60 & - & - & - & 73.6 \\
DST-as-Prompting & 56.66 & - & 57.60 & - & 83.3 & 90.6 & - \\
Seq2seq-DU & 56.10 & - & 54.40 & - & - & - & \begin{tabular}[c]{@{}c@{}}\textbf{85.0}\end{tabular} \\
DSS-DST & 60.73 & 98.05 & 58.04 & 97.66 & - & - & - \\ \hline
DiCoS-DST ($k=1$) & \begin{tabular}[c]{@{}c@{}}60.89 \\ ($\pm$0.47)\end{tabular} & \begin{tabular}[c]{@{}c@{}}98.05 \\ ($\pm$0.02)\end{tabular} & \begin{tabular}[c]{@{}c@{}}61.04 \\ ($\pm$0.56)\end{tabular} & \begin{tabular}[c]{@{}c@{}}98.05 \\ ($\pm$0.04)\end{tabular} & \begin{tabular}[c]{@{}c@{}}84.5 \\ ($\pm$1.2)\end{tabular} & \begin{tabular}[c]{@{}c@{}}91.2 \\ ($\pm$0.3)\end{tabular} & \begin{tabular}[c]{@{}c@{}}77.7 \\ ($\pm$0.2)\end{tabular} \\
DiCoS-DST ($k=2$)                  & \begin{tabular}[c]{@{}c@{}}\textbf{61.02} \\ ($\pm$0.41)\end{tabular}                                                & \begin{tabular}[c]{@{}c@{}}\textbf{98.05} \\ ($\pm$0.02)\end{tabular}                                               & \begin{tabular}[c]{@{}c@{}}\textbf{61.13} \\ ($\pm$0.54)\end{tabular}                                                & \begin{tabular}[c]{@{}c@{}}\textbf{98.06} \\ ($\pm$0.03)\end{tabular}                                               & \begin{tabular}[c]{@{}c@{}}\textbf{84.7} \\ ($\pm$1.1)\end{tabular}                                                & \begin{tabular}[c]{@{}c@{}}\textbf{91.5} \\ ($\pm$0.3)\end{tabular}              & \begin{tabular}[c]{@{}c@{}}78.4 \\ ($\pm$0.2)\end{tabular} \\
DiCoS-DST ($k=3$) & \begin{tabular}[c]{@{}c@{}}60.85 \\ ($\pm$0.24)\end{tabular} & \begin{tabular}[c]{@{}c@{}}98.05 \\ ($\pm$0.01)\end{tabular} & \begin{tabular}[c]{@{}c@{}}60.88 \\ ($\pm$0.33)\end{tabular} & \begin{tabular}[c]{@{}c@{}}98.05 \\ ($\pm$0.03)\end{tabular} & \begin{tabular}[c]{@{}c@{}}83.8 \\ ($\pm$1.1)\end{tabular} & \begin{tabular}[c]{@{}c@{}}91.0 \\ ($\pm$0.2)\end{tabular} & \begin{tabular}[c]{@{}c@{}}77.3 \\ ($\pm$0.2)\end{tabular} \\ \hline
\end{tabular}
\caption{Accuracy (\%) on the test sets of benchmark datasets vs. various approaches as reported in the literature.}
\label{table:table_1}
\end{table*}

\subsection{State Generator}

The selected dialogue content will be utilized to jointly update the dialogue state.

\paragraph{Cascaded Context Refinement}
After acquiring a nearly noise-free set $\bm{\mathrm{U}}_D$ of selected dialogue turns, we consider that directly using their representations as inputs may ignore the cross attention between them since they are used as a whole. As a result, we concatenate these dialogue utterances together to form a new input sequence $C=[\mathrm{CLS}]\oplus B_{T-1}\oplus \langle t\rangle_1\oplus D_1\oplus\ldots\oplus \langle t\rangle_{T\_S}\oplus D_{T\_S}\oplus \langle t\rangle_{T}\oplus D_{T}\ (T\_S=|\bm{\mathrm{U}}_D|)$.

Especially, we inject an indicator token ``$\langle t\rangle$'' before each turn of dialogue utterance to get aggregated turn embeddings for the subsequent classification-based state prediction. Then we feed this sequence into a single PrLM to obtain the contextualized output representation.

\paragraph{Slot Value Generation}
We first attempt to obtain the value using the extractive method from representation $C_\mathrm{E}=D_1\oplus D_2\oplus\ldots\oplus D_{T\_S}\oplus D_T$:
\begin{gather}
    p=\mathrm{softmax}(W_s C_\mathrm{E}([\mathrm{SLOT}]_{T-1}^j)^\intercal) \\
    q=\mathrm{softmax}(W_e C_\mathrm{E}([\mathrm{SLOT}]_{T-1}^j)^\intercal)
\end{gather}

The position of the maximum value in $p$ and $q$ will be the start and end predictions of the slot value. If this prediction does not belong to the candidate value set of $S_j$, we use the representation of $C_\mathrm{C}=\langle t\rangle_1\oplus \langle t\rangle_2\oplus\ldots\oplus \langle t\rangle_{T\_S}\oplus \langle t\rangle_{T}$ to get the distribution and choose the candidate slot value corresponding to the maximum value:
\begin{equation}
    y=\mathrm{softmax}(W_\mathrm{C}C_\mathrm{C}([\mathrm{SLOT}]_{T-1}^j)^\intercal)
\end{equation}

We define the training objectives of two methods as cross-entropy loss:
\begin{gather}
    L_\mathrm{ext}=-\frac{1}{|\bm{\mathrm{U}}_s|}\sum_{j}^{|\bm{\mathrm{U}}_s|}(p\log\hat{p}+q\log\hat{q}) \\
    L_\mathrm{cls}=-\frac{1}{|\bm{\mathrm{U}}_s|}\sum_{j}^{|\bm{\mathrm{U}}_s|}y\log\hat{y}
\end{gather}
where $\hat{p}$ and $\hat{q}$ are the targets indicating the proportion of all possible start and end, and $\hat{y}$ is the target indicating the probability of candidate values.

\section{Experiments}

\subsection{Datasets and Metrics}

We conduct experiments on most of the mainstream benchmark datasets on task-oriented dialogue, including MultiWOZ 2.1, MultiWOZ 2.2, Sim-R, Sim-M, and DSTC2. MultiWOZ 2.1 and MultiWOZ 2.2 are two versions of a large-scale multi-domain task-oriented dialogue dataset. It is a fully-labeled collection of human-human written dialogues spanning over multiple domains and topics. Sim-M and Sim-R are multi-turn dialogue datasets in the movie and restaurant domains, respectively. DSTC2 is collected in the restaurant domain.

We use joint goal accuracy and slot accuracy as evaluation metrics. Joint goal accuracy refers to the accuracy of the dialogue state in each turn. Slot accuracy only considers slot-level accuracy.

\subsection{Baseline Models}

We compare the performance of DiCoS-DST with the following baselines: TRADE encodes the dialogue and decodes the value using a copy-augmented decoder~\cite{wu2019transferable}. BERT-DST generates language representations suitable for scalable DST~\cite{chao2019bert}. DST+LU presents an approach for multi-task learning of language understanding and DST~\cite{rastogi2018multi}. TripPy extracts values from the dialogue context by three copy mechanisms~\cite{heck2020trippy}. DSS-DST consists of the slot selector based on the current turn dialogue, and the slot value generator based on the dialogue history~\cite{guo2021dual}. Seq2Seq-DU employs two BERT-based encoders to respectively encode the utterances and the descriptions of schemas~\cite{feng-etal-2021-sequence}. Pegasus-DST applies a span prediction-based pre-training objective designed for text summarization to DST~\cite{zhao2021effective}. DST-as-Prompting uses schema-driven prompting to provide task-aware history encoding~\cite{lee2021dialogue}.

\subsection{Implementation Details}

We employ a pre-trained ALBERT-large-uncased model~\cite{lan2019albert} for the encoder. The hidden size of the encoder $d$ is 1024. We use AdamW optimizer~\cite{loshchilov2018fixing} and set the warmup proportion to 0.01 and L2 weight decay of 0.01. We set the peak learning rate of State Update Predictor the same as in DSS-DST and the peak learning rate of the other modules to 0.0001. We set the dropout~\cite{srivastava2014dropout} rate to 0.1. We utilize word dropout~\cite{bowman2016generating} with the probability of 0.1. We set $L$ to 3. The max sequence length for all inputs is fixed to 256. During training the Multi-Perspective Dialogue Collaborative Selector, we use the ground truth selected slots instead of the predicted ones. We report the mean joint goal accuracy over 10 different random seeds to reduce statistical errors.

\begin{table}[t]
\centering
\begin{tabular}{ll}
\hline
\multicolumn{1}{c}{\begin{tabular}[c]{@{}c@{}} PrLM\end{tabular}} & MultiWOZ 2.2   \\ \hline
ALBERT (large)                                                              & \textbf{61.13} \\
ALBERT (base)                                                         & 60.05(-1.08)  \\
BERT (large)                                                        & 60.16(-0.97)  \\
BERT (base)                                                          & 59.51(-1.62)  \\ \hline
\end{tabular}
\caption{Ablation study with joint goal accuracy (\%).}
\label{table:table_2}
\end{table}

\begin{table}[t]
\centering
\begin{tabular}{ll}
\hline
\multicolumn{1}{c}{\begin{tabular}[c]{@{}c@{}} Model\end{tabular}} & MultiWOZ 2.2   \\ \hline
DiCoS-DST                                                          & \textbf{61.13} \\
-State Update Predictor                                                 & 58.48 (-2.65)  \\
\begin{tabular}[l]{@{}l@{}}-Multi-Perspective Dialogue \\\ Collaborative Selector\end{tabular}                      & 54.94 (-6.19)  \\
-Cascaded Context  Refinement                                           & 59.75 (-1.38)  \\ \hline
\end{tabular}
\caption{Ablation study with joint goal accuracy (\%). Each performance in this table represents the test results after the model was retrained with the corresponding module removed. "- State Update Predictor" means that all slots are updated in each turn. "-Multi-Perspective Dialogue Collaborative Selector" means that using the entire dialogue history without selection. "-Cascaded Context Refinement" means that directly using the representation of selected turns from the dialogue selector without context refinement.}
\label{table:table_3}
\end{table}

\subsection{Main Results}

Table~\ref{table:table_1} shows the performance of our DiCoS-DST and other baselines. Our model achieves state-of-the-art performance on MultiWOZ 2.1 and MultiWOZ 2.2 with joint goal accuracy of 61.02\% and 61.13\%. In particular, the joint goal accuracy on MultiWOZ 2.2 outperforms the previous state-of-the-art by 3.09\%. Besides, despite the sparsity of experimental results on Sim-M and Sim-R, our model still achieves state-of-the-art performance on these two datasets. On DSTC2, the performance of our model is also competitive. Among our models, DiCoS-DST ($k=2$) performs the best on all datasets. Especially, DiCoS-DST ($k=2$) and DiCoS-DST ($k=1$) perform better than DiCoS-DST ($k=3$). We conjecture that selecting two turns from the dialogue history may be sufficient, and introducing more turns may confuse the model.

\begin{table}[t]
\centering
\begin{tabular}{ll}
\hline
\multicolumn{1}{c}{\begin{tabular}[c]{@{}c@{}} Perspective(s)\end{tabular}} & MultiWOZ 2.2   \\ \hline
SN-DH & 57.73 (-3.40) \\
CT-DH & 55.47 (-5.66) \\
IMOR & 55.11 (-6.02) \\
SN-DH + CT-DH & 59.56 (-1.57) \\
SN-DH + IMOR & 58.68 (-2.45) \\
CT-DH + IMOR & 56.79 (-4.34) \\
SN-DH + CT-DH + IMOR & \textbf{61.13} \\ \hline
\end{tabular}
\caption{Ablation study with joint goal accuracy (\%). IMOR stands for Implicit Mention Oriented Reasoning.}
\label{table:table_4}
\end{table}

\begin{table}[t]
\centering
\begin{tabular}{ll}
\hline
\multicolumn{1}{c}{\begin{tabular}[c]{@{}c@{}} Graph\end{tabular}} & MultiWOZ 2.2   \\ \hline
Original Graph (DiCoS-DST) & \textbf{61.13} \\
-1st type of edges & 59.70 (-1.43) \\
-2nd type of edges & 59.62 (-1.51) \\
-3rd type of edges & 59.78 (-1.35) \\
-4th type of edges & 60.65 (-0.48) \\
\begin{tabular}[l]{@{}l@{}}+fully connecting all \\\ \ dialogue nodes\end{tabular} & 61.01 (-0.12) \\
\begin{tabular}[l]{@{}l@{}}+3rd type of edges between \\\ \ each $N_{S-V}^z$ and all $N_D^{t}$\end{tabular} & 60.04 (-1.09) \\ \hline
\end{tabular}
\caption{Ablation study with joint goal accuracy (\%).}
\label{table:table_5}
\end{table}

\subsection{Ablation Study}

\paragraph{Different PrLMs}
We employ different pre-trained language models with different scales as the backbone for training and testing on MultiWOZ 2.2. Table~\ref{table:table_2} shows that the joint goal accuracy of other encoders decreases in varying degrees compared with ALBERT (large). The joint goal accuracy of BERT(base) decreases by 1.62\%, but still outperforms the previous state-of-the-art performance on MultiWOZ 2.2. This demonstrates that our model achieves consistent performance gain in all fair comparison environments with other methods.

\paragraph{Effect of Core Components}
To explore the effectiveness of core components, we conduct an ablation study of them on MultiWOZ 2.2. As shown in Table~\ref{table:table_3}, we observe that the performance degrades by 2.65\% for joint goal accuracy when the State Update Predictor is removed. It is worth mentioning that this performance still outperforms the previous state-of-the-art performance, which demonstrates that the large performance gain of DiCoS-DST over other baselines comes from its dialogue selection. This is also supported by the observation that the performance of the model without the Multi-Perspective Dialogue Collaborative Selection module drops drastically (degrades by 6.19\% for joint goal accuracy). In addition, when we remove the Cascaded Context Refinement module, we lose 1.38\%, indicating the usefulness of interaction between different dialogue turns.

\paragraph{Separate Perspective and Combinations}
We explore the performance of each separate perspective and their various combinations. When a perspective needs to be masked, we set their corresponding gating weights to 0. It can be observed in Table~\ref{table:table_4} that the SN-DH module has the greatest impact on performance, and the most effective combination of perspectives is the combination of SN-DH and CT-DH. Despite the simplicity of the mechanism of SN-DH, the association with the slot name straightforward reflects the importance of the dialogue. To solve the common problem of coreferences in complex dialogues, the Implicit Mention Oriented Reasoning module improves the performance close enough to the CT-DH.

\paragraph{Graph Edges Ablation}
We investigate the effect of the different edges in the GNN. As shown in Table~\ref{table:table_5}, the performance degradation is relatively obvious when the first, second, and third types of edges are removed separately. It indicates that the majority of the connections are indeed to construct the reasoning logic, while the correlation of the same domain’s slots plays an auxiliary role. In addition, we design two comparative experiments. First, we start naively by fully connecting all dialogue nodes to enhance the interaction among dialogue turns. However, this change does not give a clear benefit. This is mostly because the initialization of the dialogue nodes using the dialogue representation output by MHSA already includes the contextual interactions between the dialogues. Second, we add a third type of edges between each slot-value pair node and all dialogue nodes without distinguishing the correspondence. We observe that this change does harm to the performance (degrades by 1.09\%). This reflects the importance of using other slots to explore their corresponding turns of dialogues when dealing with coreferences.

\begin{table}[t]
\centering
\begin{tabular}{ccc}
\hline
\multicolumn{3}{c}{MultiWOZ 2.2} \\ \hline
$k$ & DiCoS-DST & \begin{tabular}[c]{@{}c@{}}Granularity-Based\end{tabular}\\ \hline
1 & \textbf{61.04} & 59.58 (-1.46) \\ \hline
2 & \textbf{61.13} & 59.88 (-1.25) \\ \hline
3 & \textbf{60.88} & 59.91 (-0.97) \\ \hline
\end{tabular}
\caption{The joint goal accuracy (\%) of different $k$. The state generator is re-trained with the corresponding selected turns as input for granularity-based methods.}
\label{table:table_6}
\end{table}

\begin{table}[t]
\centering
\begin{tabular}{cccc}
\hline
\multicolumn{4}{c}{MultiWOZ 2.2} \\ \hline
Domain     & $k=0$ & $k=1$ & $k=2$ \\ \hline
Attraction & \textbf{79.15} & 79.04 & 78.79               \\
Hotel      & 56.95 & \textbf{58.07} & 58.02               \\
Restaurant & 73.81 & 74.73 & \textbf{75.14}               \\
Taxi       & 53.50 & 55.12 & \textbf{56.33}               \\
Train      & 75.13 & 76.89 & \textbf{77.26}               \\ \hline
\end{tabular}
\caption{Domain-specific results on MultiWOZ 2.2.}
\label{table:table_7}
\end{table}

\section{Analysis}

\subsection{Is It Beyond the Granularity?}

DiCoS-DST filters out some distracting information by selecting relevant dialogues, but is it really beyond the granularity? To investigate it, we simulate the granularity and compare it with DiCoS-DST. Specifically, we use the maximum granularity (i.e., the number of dialogue turns spanning from the selected furthest dialogue turn to the current turn) and capture the corresponding dialogue contents as input to State Generator. As shown in Table~\ref{table:table_6}, DiCoS-DST outperforms the granularity-based method by 1.46\% ($k=1$), 1.25\% ($k=2$), and 0.97\% ($k=3$), indicating that there is still redundant information in the dialogue contents determined by the granularity that confuses the model.




\subsection{Domain-Specific Dialogue Requirements}

Table~\ref{table:table_7} shows the domain-specific results when we set different values for $k\ (k=0, 1, 2)$. In $taxi$ and $train$ domains, the performance of the model decreases significantly when $k=0$ compared to $k=2$, implying that acquiring the values of the slots in these domains is highly dependent on the dialogue history. Nevertheless, there is no significant difference in the performance in $attraction$ domain when we set different values for $k$. This indicates that the values of the slots in this domain can usually be simply obtained from the current turn dialogue, instead of using the dialogue history or resolving coreferences.

\section{Conclusion}
We introduce an effective DiCoS-DST that dynamically selects the relevant dialogue contents corresponding to each slot from a combination of three perspectives. The dialogue collaborative selector module performs a comprehensive selection for each turn dialogue based on its relation to the slot name, its connection to the current turn dialogue, and the implicit mention oriented reasoning. Then only the selected dialogue contents are fed into State Generator, which explicitly minimizes the distracting information passed to the downstream state prediction. Our DiCoS-DST model achieves new state-of-the-art performance on the MultiWOZ benchmark, and achieves competitive performance on most other DST benchmark datasets. The potential relationship among the above perspectives is a promising research direction, and we will explore it for more than dialogue selection in the future.

\section*{Acknowledgements}
This work was supported by Beijing Natural Science Foundation(Grant No. 4222032) and BUPT Excellent Ph.D. Students Foundation. We thank the anonymous reviewers for their insightful comments.


\section*{Ethical Considerations}
The claims in this paper match the experimental results. This work focuses on DST in task-oriented dialogue systems, and the improvements could have a positive impact on helping humans to complete goals more effectively in a more intelligent way of communication.

\bibliography{anthology,custom}
\bibliographystyle{acl_natbib}




\end{document}